\documentclass[10pt,twocolumn,letterpaper]{article}

\usepackage{iccv}
\usepackage{times}
\usepackage{epsfig}
\usepackage{graphicx}
\usepackage{amsmath}
\usepackage{amssymb}
\usepackage{multirow} 
\usepackage{makecell}
\usepackage{booktabs}

\usepackage[pagebackref=true,breaklinks=true,letterpaper=true,colorlinks,bookmarks=false]{hyperref}

\iccvfinalcopy 

\begin{document}

\title{FreeCOS: Self-Supervised Learning from Fractals and Unlabeled Images for Curvilinear Object Segmentation}

\author{
Tianyi Shi, Xiaohuan Ding,  Liang Zhang, Xin Yang\footnotemark[2]\\
School of EIC, Huazhong University of Science \& Technology\\
{\tt\small \{shitianyihust, dingxiaohuan, liangz, xinyang2014\}@hust.edu.cn}
}

\maketitle

\begin{abstract}
Curvilinear object segmentation is critical for many applications. However, manually annotating curvilinear objects is very time-consuming and error-prone, yielding insufficiently available annotated datasets for existing supervised methods and domain adaptation methods. This paper proposes a self-supervised curvilinear object segmentation method that learns robust and distinctive features from fractals and unlabeled images (FreeCOS). The key contributions include a novel Fractal-FDA synthesis (FFS) module and a geometric information alignment (GIA) approach. FFS generates curvilinear structures based on the parametric Fractal L-system and integrates the generated structures into unlabeled images to obtain synthetic training images via Fourier Domain Adaptation. GIA reduces the intensity differences between the synthetic and unlabeled images by comparing the intensity order of a given pixel to the values of its nearby neighbors. Such image alignment can explicitly remove the dependency on absolute intensity values and enhance the inherent geometric characteristics which are common in both synthetic and real images. In addition, GIA aligns features of synthetic and real images via the prediction space adaptation loss (PSAL) and the curvilinear mask contrastive loss (CMCL). 
Extensive experimental results on four public datasets, i.e., XCAD, DRIVE, STARE and CrackTree demonstrate that our method outperforms the state-of-the-art unsupervised methods, self-supervised methods and traditional methods by a large margin. The source code of this work is available at https://github.com/TY-Shi/FreeCOS.
\end{abstract}

\section{Introduction}
Automatically segmenting curvilinear structures (such as vascular trees in medical images and road systems in aerial photography) is critical for many applications, including retinal fundus disease screening~\cite{abramoff2010retinal,fraz2012blood}, diagnosing coronary artery disease~\cite{thomas2017novel}, road condition evaluation and maintenance~\cite{zou2012cracktree}. Despite a plethora of research works in the literature, accurately segmenting curvilinear objects remains challenging due to their complex structures with numerous tiny branches, tortuosity shapes, ambiguous boundaries due to imaging issues and noisy backgrounds.

Most recent methods~\cite{wang2019context,cheng2021joint,shit2021cldice,wang2020deep,hu2019topology,mosinska2018beyond,batra2019improved,mosinska2019joint,bastani2018roadtracer,shi2022local} leverage supervised deep learning for curvilinear object segmentation and have achieved encouraging results. However, those methods require a large number of pixel-wise manual annotations for training which are very expensive to obtain and error-prone due to poor image quality, annotator’s fatigue and lack of experience. Although, there are several publically available annotated datasets for curvilinear object segmentation~\cite{staal2004ridge,hoover2000locating,zou2012cracktree,ma2021self}, the large appearance variations between different curvilinear object images, e.g., X-ray coronary angiography images vs. retinal fundus images, yields significant performance degradation for supervised models across different types of images (even across the same type of images acquired using different equipments). As a result, expensive manual annotations are inevitably demanded to tune the segmentation model for a particular application. Potential solutions to alleviate the annotation burden include domain adaption ~\cite{bermudez2018domain,roels2019domain} and unsupervised segmentation~\cite{chen2019unsupervised,ji2019invariant,li2021contrastive, do2021clustering,chen2019unsupervised,abdal2021labels4free}. However, the effectiveness of domain adaptation is largely dependent on the quality of annotated data in the source domain and constrained by the gap between the source and target domain. Existing unsupervised segmentation methods~\cite{chen2019unsupervised,ji2019invariant} can hardly achieve satisfactory performance for curvilinear objects due to their thin, long, and tortuosity shapes, complex branching structures, and confusing background artifacts. 

Despite the high complexity and great variety of curvilinear structures in different applications, they share some common characteristics (i.e., the tube-like shape and the branching structure). Thus, existing studies~\cite{zamir2001arterial,zamir1999fractal,zamir2001fractal} have demonstrated that several curvilinear structures (e.g., arterial trees of the circulation system) can be generated via the fractal systems with proper branching parameters to mimic the fractal and physiological characteristics, and some observed variability. These results motivate us to use the generated curvilinear objects via the fractal systems to explicitly encode geometric properties and varieties (i.e., different diameters and lengths of branches, and different branching angles) into training samples and to assist feature learning of a curvilinear structure segmentation model. 
However, such formulas generated training samples can hardly mimic the appearance patterns within curvilinear objects, the transition regions between curvilinear objects and backgrounds, which are also key information for learning a segmentation model and contained in easily-obtained unlabeled target images. This paper asks the question, how to combine fractals and unlabeled target images to encode sufficient and comprehensive visual cues for learning robust and distinctive features of curvilinear structures? 

The main contribution of this paper is a self-supervised segmentation method based on a novel Fractal-FDA synthesis (FFS) module and a geometric information alignment (GIA) approach. Specifically, curvilinear structures are synthesized by the parametric fractal L-Systems~\cite{zamir2001arterial} and serve as segmentation labels of synthetic training samples. To simulate appearance patterns in the object-background transition regions and background regions, we apply Fourier Domain Adaptation~\cite{yang2020fda} (FDA) to fuse synthetic curvilinear structures and unlabeled target images. The synthetic images via our FFS module can effectively guide learning distinctive features to distinguish curvilinear objects and backgrounds. To further improve the robustness to differences between intensity distributions of synthetic and real target images, we design a novel geometric information alignment (GIA) approach which aligns information of synthetic and target images at both image and feature levels. Specifically, GIA first converts each training image (synthetic and target images) into four geometry-enhanced images by comparing the intensity order of a given pixel to the values of its nearby neighbors (i.e., along with the up, down, left and right directions). In this way, the four converted images do not depend on the absolute intensity values but the relative intensity in order to capture the inherent geometric characteristic of the curvilinear structure, reducing the intensity differences between  synthetic and target images. Then, we extract features from the 4-channel converted images and propose two loss functions, i.e., a prediction space adaptation loss (PSAL) and a curvilinear mask contrastive loss (CMCL), to align the geometric features of synthetic and target images. The PSAL minimizes the distance between the segmentation masks of the target images and synthetic curvilinear objects and the CMCL minimizes the distance between features of segmented masks and synthetic objects.

The FreeCOS based on FFS and GIA approaches applies to several public curvilinear object datasets, including XCAD~\cite{ma2021self}, DRIVE~\cite{staal2004ridge}, STARE~\cite{hoover2000locating} and CrackTree~\cite{zou2012cracktree}. Extensive experimental results demonstrate that FreeCOS outperforms the state-of-the-art self-supervised~\cite{ma2021self,kim2022diffusion}, unsupervised~\cite{chen2019unsupervised,ji2019invariant}, and traditional methods~\cite{frangi1998multiscale,law2008three}. To summarize, the main contributions of this work are as follows:

\begin{itemize}
     \item We propose a novel self-supervised curvilinear feature learning method which intelligently combines tree-like fractals and unlabeled images to assist in learning robust and distinctive feature representations.
     \item We propose Fractal-FDA synthesis (FFS) and geometric information alignment (GIA), which are the two key enabling modules of our method. FFS integrates the synthetic curvilinear structures into unlabeled images to guide learning distinctive features to distinguish foregrounds and backgrounds. GIA enhances geometric features and meanwhile improves the feature robustness to intensity differences between synthetic and target unlabeled images.
     \item We develop a novel self-supervised segmentation network that can be trained using only target images and fractal synthetic curvilinear objects. Our network performs significantly better than state-of-the-art self-supervised /unsupervised methods on multiple public datasets with various curvilinear objects.
\end{itemize}

\begin{figure*}
\begin{center}
\includegraphics[width=1\linewidth]{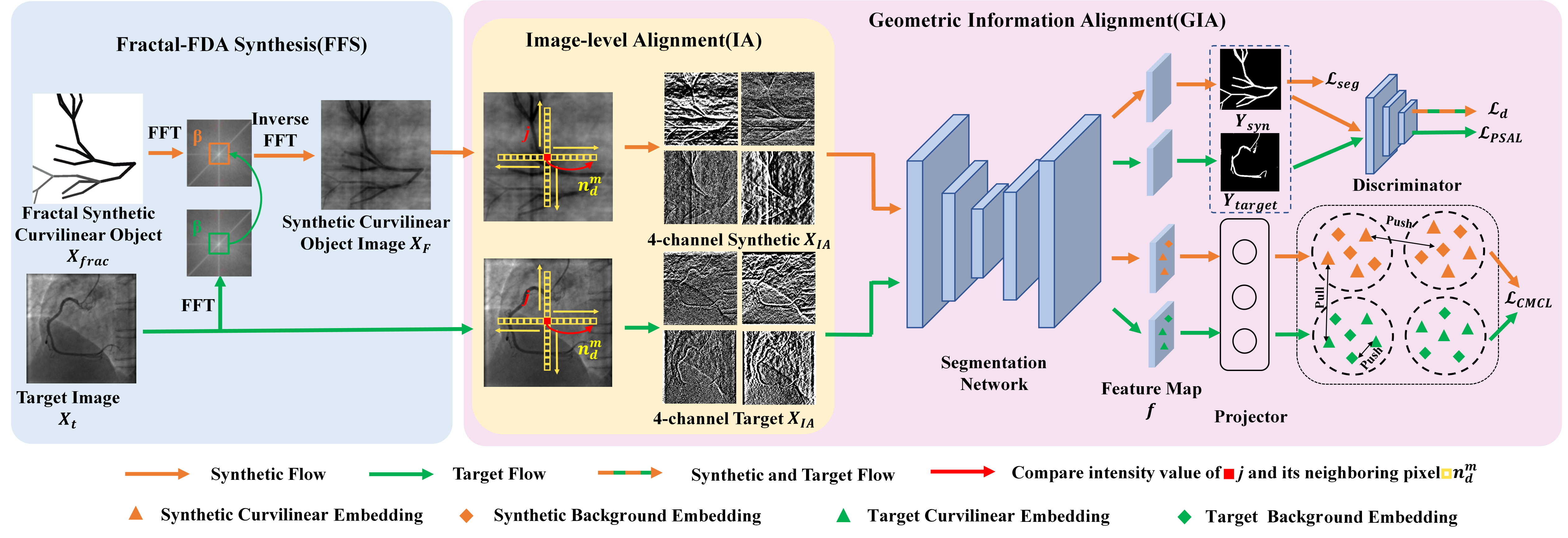}
\end{center}
\vspace{-10pt}
\caption{The pipeline of Self-Supervised Learning from Fractals and Unlabeled Images for Curvilinear Object Segmentation (FreeCOS).}
\label{fig:pipeline}
\end{figure*}

\section{Related Work}
\label{sec:related}
\subsection{Traditional Methods}
Traditional curvilinear object segmentation methods~\cite{khan2020hybrid,memari2019retinal,frangi1998multiscale,wang2020higher,law2008three} design heuristic rules and/or filters to capture features of the target curvilinear objects. 
For instance, Frangi et al. introduce the vesselness filter~\cite{frangi1998multiscale} based on the Hessian matrix to represent and enhance tube-like curvilinear objects.
Khan et al.~\cite{khan2020hybrid} further design B-COSFIRE filters to denoise retinal images and segment retinal vessels.
Memari et al.~\cite{memari2019retinal} enhance image contrast via contrast-limited adaptive histogram equalization and then segment retinal vessels based on hand-crafted filters.
In~\cite{wang2020higher,law2008three}, the authors propose optimally oriented flux (OOF) to enhance curvilinear tube-like objects. OOF exhibits better performance for segmenting adjacent curvilinear objects yet is sensitive to different sizes of curvilinear objects.

Traditional methods based on hand-crafted filters do not require any training yet they require careful parameter tuning for optimized performance. And the optimized parameter settings are usually data-dependent or even region-dependent, limiting their convenience in segmenting a wide variety of curvilinear objects.

\subsection{Unsupervised Segmentation Methods}
Unsupervised segmentation methods can be generally divided into two classes: clustering based~\cite{ji2019invariant, li2021contrastive, do2021clustering} and adversarial learning based~\cite{chen2019unsupervised,abdal2021labels4free}.
Xu et al.~\cite{ji2019invariant} propose Invariant Information Clustering (IIC) which automatically partitions input images into regions of different semantic classes by optimizing mutual information between related region pairs. 
Such a clustering-based method is more suitable for segmenting objects with aspect ratios close to one while becoming ineffective for curvilinear objects due to their thin, long, tortuous shapes.
Redo~\cite{chen2019unsupervised} is based on an adversarial architecture where the generator is guided by an image and extracts the object mask, then redraws a new object at the same location with different textures/ colors.
However, this adversarial learning-based unsupervised method only performs well for objects which are visually distinguishable from backgrounds. For the segmentation of curvilinear objects with complex and numerous tiny branching structures, embedded in confusing and cluttered backgrounds, the efficacy of such a method degrades significantly. 

In contrast to unsupervised methods, our method explicitly encodes geometric and photometric characteristics, as well as some observed varieties of curvilinear objects in target application into synthetic images. Those synthetic images provide labels to effectively guide the model to learn robust and distinctive features and thus yield superior performance to state-of-the-art unsupervised methods~\cite{chen2019unsupervised,ji2019invariant}.

\subsection{Self-supervised Learning Methods}
Self-supervised learning methods construct pretexts from large-scale unsupervised data and utilize contrastive learning losses to measure the similarities of sample pairs in the representation space. To this end, various pretext tasks have been designed, including jigsaw~\cite{doersch2015unsupervised}, hole-fill~\cite{pathak2016context} and transformation invariance~\cite{gidaris2018unsupervised}. Although existing self-supervised methods have achieved outstanding performance in classification~\cite{misra2020self,azizi2021big}, detection~\cite{xie2021detco,bar2022detreg}, and image translation~\cite{park2020contrastive,wu2021contrastive}, few of them provide a suitable pretext task for curvilinear object segmentation.
A potential solution is the pixel-level contrastive-based method~\cite{wang2022contrastmask,alonso2021semi,zhong2021pixel}, while it requires heavy manual annotations to prepare pixel-level positive and negative samples.
Similar work to ours is~\cite{ma2021self,kim2022diffusion} which designs a self-supervised vessel segmentation method via adversarial learning and fractals. But this method requires clean background images as input (i.e., the first frame of the angiography sequence) for synthesis which greatly limits its applications. In addition, adversarial learning cannot explicitly and precisely enforce visual cues, which are important for learning segmentation-oriented features, being encoded in the synthetic images. As a result,  although  their synthetic images visually look similar to the target images, the segmentation accuracy is also quite low. 

\section{Method}
\label{sec:method}
Figure.~\ref{fig:pipeline} shows the framework of FreeCOS which consists of two main modules, i.e., Fractal-FDA Synthesis (FFS) and Geometric Information Alignment (GIA). In FFS, we generate synthetic curvilinear structures via the parametric Fractal L-Systems and use the generated structures as segmentation maps to guide self-supervised training of a segmentation network. The synthetic curvilinear structures are then integrated into unlabeled images via FDA to form synthetic images of curvilinear structures. The intensity distributions of synthetic images could deviate from those of real target images and hence yield poor feature robustness. To address this problem, our GIA module first reduces image-level differences between the synthetic and target images by converting intensity images into four-channel intensity order images. The converted images of both synthetic images and target images are then input into U-Net to extract feature representations. Our GIA further aligns features of synthetic images and target images via the prediction space adaptation loss (PSAL) and the curvilinear mask contrastive loss (CMCL). In the following, we present the details of FFS and GIA.

\subsection{Fractal-FDA Synthesis}
\begin{figure}
\begin{center}
\includegraphics[width=1.0\linewidth]{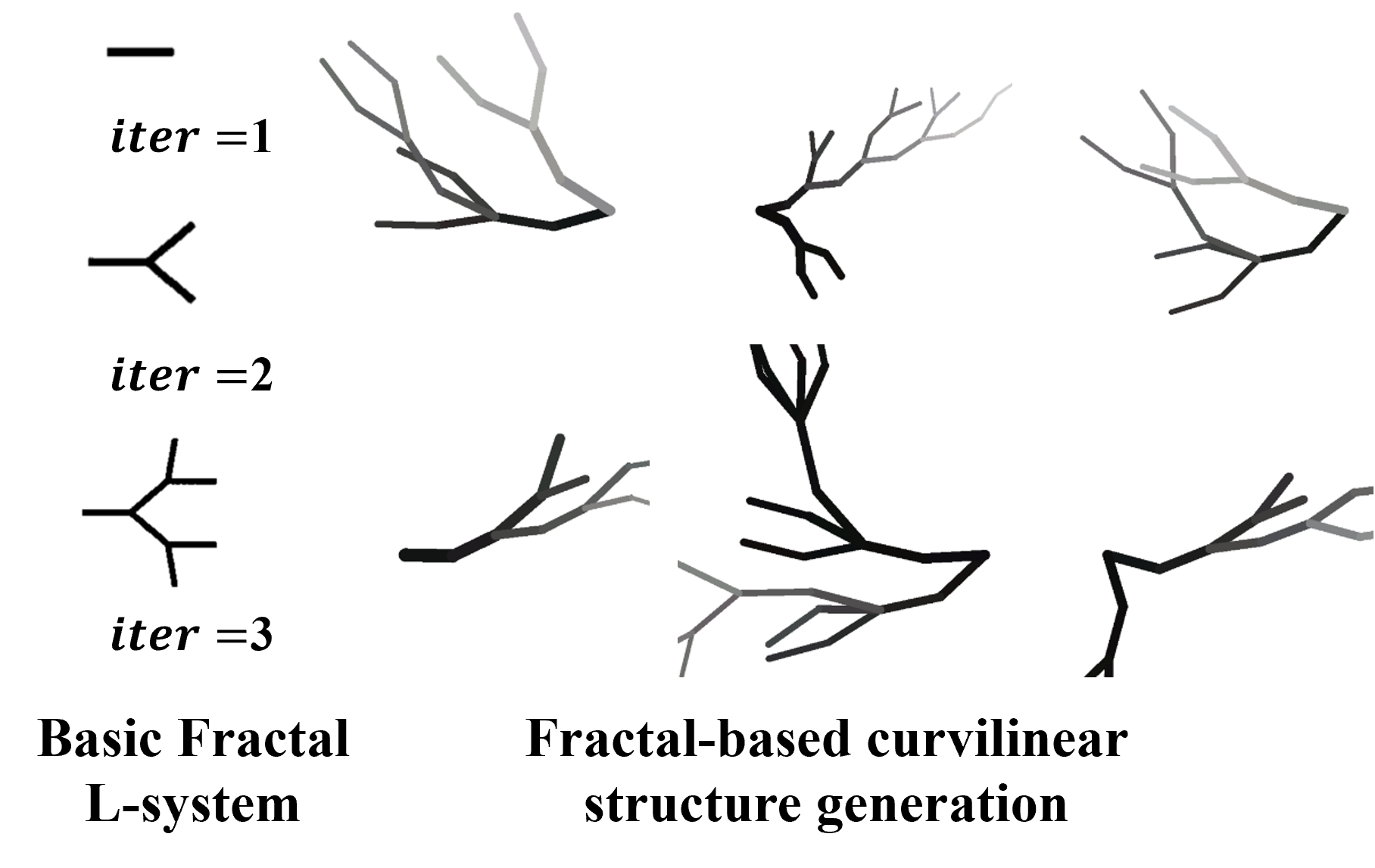}
\end{center}
\vspace{-10pt}
\caption{Exemplar results of the generated curvilinear structures from the basic and parametric Fractal L-system.}
\label{fig:fractal}
\end{figure}
We first generate curvilinear structures via the parametric Fractal L-Systems and then integrate the synthetic objects into unlabeled target images to obtain synthetic training samples via FDA.

\textbf{Fractal-based curvilinear structure generation}. 
Fractals are simple graphic patterns rendered by mathematical formulas. In this work, we adopt the parametric Fractal L-systems proposed by Zamir et al.~\cite{zamir2001arterial} to generate fractal tree structures and meanwhile select proper branching parameters according to physiological laws of curvilinear objects in target applications. Specifically, “grammar” for generating curvilinear structures with repeated bifurcations based on Fractal L-systems method is defined as follows: 
\begin{equation}
\label{eq:rule}
\begin{aligned}
\omega: & \; \quad F\\
rule: & \; \quad F \rightarrow F[-F][+F]
\end{aligned}
\end{equation}
The generated object by iteration $iter$:
\begin{equation}
\label{eq:draw_base}
\operatorname{Draw}(iter, F , rule)
\end{equation}
where $F$ represents a line of unit length in the horizontal direction, $\omega$ denotes an axiom, $iter$ denotes the iteration and $rule$ denotes the production rule. The square brackets represent the departure from ($[$) and return to ($]$) a branch point. The plus and minus signs represent turns through a given angle $\delta$ in the clockwise and anticlockwise directions, respectively. For example, the first three stages of a tree produced by the basic Fractal L-system, denoted by $iter$=1 $\sim$ 3,  are given by the:
\begin{equation}
\label{eq:iter_example}
\begin{array}{ll}
iter=1 & F \\
iter=2 & F[-F][+F] \\
iter=3 & F[-F][+F][-F[-F][+F]][+F[-F][+F]]
\end{array}
\end{equation}
The left part of Figure.~\ref{fig:fractal} illustrates exemplar results of the generated curvilinear structures based on the basic Fractal system.

To synthesize curvilinear structures with various widths and lengths, we replace $F$ using $F_{random}\left(w_i, l_i, c_i\right)$ which represents a line of width $w_i$, length $l_i$ and intensity $c_i$. We set the initial width $w_{init}$, length $l_{init}$ and decreased parameter $\gamma$ by the repeated $F_{random}$  in the $Draw$ grammar. For the index number $i$ of $F_{random}$, $w_i$ and $l_i$ are given by:
\begin{equation}
\begin{aligned}
w_i &=w_{init} \cdot \gamma^{i-1} \\
l_i &=l_{init} \cdot \gamma^{i-1}
\end{aligned}
\end{equation}

We randomly choose a value within the range of (0, 255) and set this value as the intensity $c_i$ for the index $i$ of $F_{random}$. For the branching angle $\delta$, we replace $\delta$ using $\delta_{random}$ (defined by $\delta_{init} \pm \delta_{delta}$, $\delta_{init}$ 
within the range of $\left(20^{\circ}, 120^{\circ}\right)$), in which $\delta_{delta}$ is a random angle ranging from $10^{\circ}$ to $40^{\circ}$. 

To mimic the geometric characteristics of a target application, we incorporate the above-mentioned parameters into the L-Systems. Specifically, we design a set of rules $ruleset$ : $\left(\right.$$rule_1, \cdots$, $\left.rule_n\right)$ for different fractal structures, like $\quad F_{random}\rightarrow F_{random}-F_{random}\left[+F_{random}-\right.$
$\left.F_{random}\right]\left[-F_{random}+F_{random}\right]$. For each iteration $iter$, we randomly select a rule from the $ruleset$ as the production $rule$. In this way, we can increase the diversity of structures. The generated fractal curvilinear object $X_{frac} \in \mathbb{R}^{H \times W}$ by iteration is given by:
\begin{equation}
\label{eq:Draw_random}
X_{frac}=\operatorname{Draw}(iter, F_{random} , ruleset,w_i, l_i, c_i)
\end{equation}

\textbf{FDA-based curvilinear object image synthesis}. 
We further incorporate synthetic curvilinear objects into the target unlabeled images via FDA~\cite{yang2020fda}. We define the target images as $X_{t} \in \mathbb{R}^{H \times W}$ and the synthetic images of fractal curvilinear object as $X_{frac}$. Let $\mathcal{F}^A: \in \mathbb{R}^{H \times W} \rightarrow \in \mathbb{R}^{H \times W}$ be the amplitude component of the Fourier transform $\mathcal{F}$ of a grayscale image. We define a $\beta \in(0,1)$ to select the center region of the amplitude map.

Given two randomly sampled synthetic image $X_{frac}$ and target image $X_{t}$, we follow FDA~\cite{yang2020fda} to replace the low-frequency part in the amplitude map of $X_{frac}$ by the FFT~\cite{frigo1998fftw}, denotes as $\mathcal{F}^A\left(X_{frac}\right)$, with that of the target image $X_{t}$, denoted as $\mathcal{F}^A\left(X_{t}\right)$. Then, the modified spectral representation of $X_{frac}$, with its phase component unaltered, is mapped back to image $X_{frac \rightarrow t}$ by the inverse FFT~\cite{frigo1998fftw}, whose content is the same as $X_{frac}$, but will resemble the appearance of a sample $X_t$.
After that, we apply a Gaussian blur to the output synthetic image of FDA to obtain synthetic images from FFS, denoted as $X_{F}$.

\subsection{Geometric Information Alignment}
Synthetic images generated by FFS could still have nontrivial intensity differences from that of real images. To address this problem, we further align synthetic images and real target images at both image and feature levels.

\textbf{Image-level Alignment}.
We aim to explicitly remove the dependency on raw intensity values for both synthetic and real images and meanwhile enhance the geometric characteristics of curvilinear structures. 
To this end, for each image pixel $j$ we compare its intensity value $X(j)$ with its 8 neighboring pixels $\{n_d^m \, | \, m = 1, \dots, 8\}$ (yellow box in Figure.~\ref{fig:pipeline}) lying perpendicular to $j$ (red box in Figure.~\ref{fig:pipeline}), where directional $d$ denotes left, right, top, and bottom side of $j$. If $X(j)$ $\textgreater$ $X\left(n_d^m\right)$, we set the corresponding value for $n_d^m$ to 0. Otherwise, we set the value for $n_d^m$ to 1. Such operation produces four 8-bit images for each input image, where the $m$-th bit on the $d$-th image denotes the relative intensity order between $j$ and its neighbor $m$ along the $d$-th direction. For each input image $X$ (denote as $X_{F}$ and $X_{t}$), we concatenate four 8-bit images $\left(X_{IA}^1, \cdots, X_{IA}^d\right)$, where $X_{IA}^d(j)$ is represented as
\begin{equation}
\label{eq:ICA}
X_{IA}^d(j)=\sum_{m=1}^8\left[X(j)>X\left(n_d^m\right)\right] \times 2^{m-1}
\end{equation}

Such local intensity order transformation~\cite{shi2022local} can capture the intrinsic of the curvilinear object and meanwhile reduces the intensity gap between synthetic and real images. 

\textbf{Feature-level Alignment}. Given the 4-channel transformed images, we utilize U-Net~\cite{ronneberger2015u} to extract features. We align curvilinear objects’ features of synthetic images and unlabeled target images based on two loss functions, i.e., the prediction space adaptation loss (PSAL) and the curvilinear mask contrastive loss (CMCL). 

1) \textbf{Prediction space adaptation loss} We utilize adversarial learning to explicitly align the prediction space distribution of target images and synthetic images. We denote the prediction segmentation masks of the target images and the synthetic images as $Y_{target} \in \mathbb{R}^{H \times W}$ and $Y_{syn} \in \mathbb{R}^{H \times W}$, respectively. We input $Y_{target}$ and $Y_{syn}$ into a fully-convolutional discriminator $D$ as~\cite{zhou2020affinity} trained via a binary cross-entropy loss $\mathcal{L}_{d}$:
\begin{equation}
\label{eq:L_d}
\mathcal{L}_{d}=\mathbb{E}\left[\log \left(D\left(Y_{\text {syn }}\right)\right)\right]+\mathbb{E}\left[\log \left(1-D\left(Y_{target}\right)\right)\right]
\end{equation}
Accordingly, the PSAL is computed as:
\begin{equation}
\label{eq:L_adv}
\mathcal{L}_{PSAL}=\mathbb{E}\left[\log \left(D\left(Y_{target}\right)\right)\right]
\end{equation}

2) \textbf{Curvilinear mask contrastive loss}. The CMCL aims to reduce the distance between features of synthetic images and true target images, and meanwhile to improve the feature distinctiveness between curvilinear objects and backgrounds. To this end, we take the feature maps (denoted as $f \in \mathbb{R}^{H \times W \times C}$) from the final decoder layer of U-Net, the mask of synthetic image (denoted as $G_{syn}$$\in\{0,1\}^{H \times W}$) and prediction mask of target image (denoted $Y_{target}$$\in[0,1]$) as input. We process $f$ using a lightweight contrastive encoder and a contrastive projector as~\cite{wang2022contrastmask} to map $f$ to the feature space where the pixel-level contrastive loss is applied for $Z \in \mathbb{R}^{H \times W \times C}$.

We denote $I$ as ${H \times W}$ spatial location of the projected feature maps $Z$, then for a location $i \in I$, we can obtain a feature vector $z_i$ at location $i$ from feature map $Z$, label values $g_i$ and $y_i$ at $i$ from the mask of synthetic image and the prediction mask respectively. We partition pixels of $I$ into two groups: curvilinear object locations $I^+$ and background locations $I^-$. For synthetic images, we perform the partition directly based on the mask $G_{syn}$, i.e., $I_{syn}^{+}=\left\{i \in I_{syn} \mid g_{i}=1\right\}$ and $I_{syn}^{-}=\left\{i \in I_{syn} \mid g_{i}=0\right\}$. For target images, as the ground-truth mask is not available, we alternatively perform the partition based on the prediction probability mask $Y_{target}$, i.e., $I_{target}^{+}=\left\{i \in I_{target} \mid y_{i} \geq 1-\alpha\right\}$ and $I_{target}^{-}=\left\{i \in I_{target} \mid y_{i} \leq \alpha\right\}$, where $\alpha$=0.1 is a small threshold and is fixed in our method.

We let $q_{syn}^{+}=\left\{\mathbf{z}_{i} \mid i \in \S\left(I_{syn}^{+}, \sigma\right)\right\}$  and $k_{target}^{+}=\left\{\mathbf{z}_{i} \mid i \in \S\left(I_{target}^{+}, \sigma\right)\right\}$ denote the curvilinear keys of synthetic and target images respectively. Similarly, we define $k_{syn}^{-}=\left\{\mathbf{z}_{i} \mid i \in \S\left(I_{syn}^{-}, \sigma\right)\right\}$ and $k_{target}^{-}=\left\{\mathbf{z}_{i} \mid i \in \S\left(I_{target}^{-}, \sigma\right)\right\}$ as the background keys of synthetic and target images, where $\S(\bullet, \sigma)$ is a random sampling operator which samples a subset from a set randomly with a proportion ratio $\sigma$. We combine $N$ negative queries of the features of synthetic and target images to form a negative set $k^{-}=\left(k_{syn}^{-}, k_{target}^{-}\right)$. The CMCL is defined as:
\begin{small}
\begin{equation}
\begin{aligned}
\label{eq:L_con}
&\mathcal{L}_{CMCL}=-\log\left( \exp \left(q_{syn}^{+} \cdot k_{target}^{+} / \tau\right)\right)+\\
&\log \left(\exp \left(q_{syn}^{+} \cdot k_{target}^{+} / \tau\right)
+\sum_{i=0}^{N} \exp \left(q_{syn}^{+} \cdot k_{i}^{-} / \tau\right)\right)
\end{aligned}
\end{equation}
\end{small}
where $\tau$ is a temperature hyper-parameter.

3) \textbf{Final Loss}. The final loss is a combination of the segmentation loss, the PSAL and CMCL as:
\begin{equation}
\label{eq:L_seg}
\mathcal{L}_{seg}=\mathbb{E}\left[G_{syn} \cdot \log \left(Y_{syn}\right)\right]
\end{equation}
\begin{equation}
\label{eq:L_toal}
\mathcal{L}=\mathcal{L}_{seg}+\mathcal{L}_{PSAL}+\lambda \mathcal{L}_{CMCL}
\end{equation}

\section{Experiments}
\label{sec:experiments}
\textbf{XCAD dataset}. The X-ray angiography coronary artery disease (XCAD) dataset~\cite{ma2021self} is obtained during stent placement using a General Electric Innova IGS 520 system. Each image has a resolution of 512 × 512 pixels with one channel. The training set contains 1621 coronary angiograms without annotations as target images. The testing set contains 126 independent coronary angiograms with vessel segmentation maps annotated by experienced radiologists.

\textbf{Retinal dataset}. We also employ two public retinal datasets to validate the effectiveness of the proposed method. The DRIVE dataset~\cite{staal2004ridge} consists of 40 color retinal images of size 565 × 584 pixels. We use 20 images as target images and 20 remaining as test images. The STARE dataset~\cite{hoover2000locating} contains 20 color retinal images of size 700 × 605 pixels with annotations as test images. There are 377 images without annotation which are used as target images.

\textbf{CrackTree dataset}. The CrackTree dataset~\cite{zou2012cracktree} contains 206 800×600 pavement images with different kinds of cracks with curvilinear structures. The whole dataset is split into 160 target images and 46 test images by~\cite{shi2022local} setting. Following~\cite{shi2022local}, we dilate the annotated centerlines by 4 pixels to form the ground-truth segmentation.

\subsection{Evaluation Metrics}
For XCAD and CrackTree, we follow~\cite{ma2021self,shi2022local} to use the following widely-used metrics in our evaluation, i.e., Jaccard Index (Jaccard), Dice Coefficient (Dice), accuracy (Acc.), sensitivity (Sn.) and specificity (Sp.). For the DRIVE and STARE datasets, we follow the state-of-the-art works for retina vessel segmentation~\cite{ma2021self} to report accuracy (Acc.), sensitivity (Sn.) specificity (Sp.) and area under curve (AUC) in our evaluation. 

\subsection{Implementation Details}
For FFS, we set the $ruleset$ as $\left(\right.$$rule_1, \cdots$$,$ $\left.rule_4\right)$, such as:

\text{$rule_1$}:$F_{random}\rightarrow F_{random}\left[+F_{random}-F_{random}\right]$.

\text{$rule_2$}:$F_{random}\rightarrow F_{random}\left[-F_{random}-F_{random}\right]$. 

\text{$rule_3$}:$F_{random}\rightarrow F_{random}-F_{random}-F_{random}$.

\text{$rule_4$}:$F_{random}\rightarrow F_{random}+F_{random}+F_{random}$.

We set the initial parameters of the Fractal system as follows. For all four datasets, the angle is randomly selected from $\left(20^{\circ}, 120^{\circ}\right)$, the initial length is randomly selected from (120px, 200px), the decreased parameter $\gamma$ is randomly selected from (0.7, 1). The initial width $w_{init}$ is ranging from (8px, 14px) for XCAD, DRIVE and STARE. For CrackTree, the $w_{init}$ is selected from (2px, 6px). The  kernel size of Gaussian blur for FFS is 13.
We generate 150, 150, 600 synthetic fractal images $X_{frac}$ for XCAD, CrackTree, and retinal datasets, respectively.

We apply data augmentation including horizontal flipping, random brightness and contrast changes ranging from 1.0 to 2.1, random saturation ranging from 0.5 to 1.5, and random rotation with $90^{\circ}$, $180^{\circ}$, and $270^{\circ}$. The standard deviation for Gaussian noise is set to a random value within (-5, +5). All images are cropped to 256×256 pixels for training. All the data-augmented operations are applied before the GIA module. The segmentation network is trained using the SGD with a momentum of 0.9 for optimization and the initial learning rate is 0.01. The discriminator network is trained using an Adam optimizer with an initial learning rate of $10^{-3}$. We employ a batch size of 8 to train the network for 600 epochs. The number of negative queries  $q_{syn}^{+}$, $k_{target}^{+}$ and $k^{-}$ per batch is taken up to 500, 500 and 1000.
The amplitude map center region selection  parameter of $\beta$ is set to 0.3. the sampling ratio is 0.3, the hyper-parameter $\lambda$ is 0.4 and the temperature hyper-parameter $\tau$ is 0.1.
\subsection{Experimental Results}
\subsubsection{Comparison with State-of-the-art}

\begin{table}
\centering
\resizebox{1\linewidth}{!}{
\begin{tabular}{clccccc}
\toprule
& Methods  & Jaccard    & Dice      & Acc.      & Sn.      & Sp.\\
\midrule
Upper bound& U-Net~\cite{ronneberger2015u}  &0.571 &0.724 &0.981 &0.868 &0.996\\
\midrule
\multirow{3}{*}{\makecell{Domain \\ Adaptation}} & U-Net~\cite{ronneberger2015u} &0.228 &0.365 &0.831 &0.444 &0.906\\
                                          & MMD~\cite{bermudez2018domain}  &0.262 &0.416 &0.873 &0.553 &0.920\\
                                          & YNet~\cite{roels2019domain} &0.287 &0.434 &0.891 &0.523 &0.935\\ 
\midrule
\multirow{2}{*}{Traditional}  
                                          & Hessian~\cite{frangi1998multiscale} &0.307 &0.465 &0.948 &0.406 &0.981\\
                                          & OOF~\cite{law2008three}  &0.241	&0.386	&0.899	&0.566	&0.920\\
                                          \midrule
\multirow{2}{*}{Unsupervised}  
                                          & IIC~\cite{ji2019invariant}  &0.124 &0.178 &0.738 &0.487 &0.754\\
                                          & ReDO~\cite{chen2019unsupervised} &0.151 &0.261 &0.753 &0.392 &0.923\\ 
\midrule       
\multirow{3}{*}{Self-supervised}  
                                          & SSVS~\cite{ma2021self}  &0.389 &0.557 &0.945 &0.583 &0.972\\
                                          & DARL~\cite{kim2022diffusion}  &0.471 &0.636 &\textbf{0.962} &0.597 &\textbf{0.985}\\
                                          & Ours  & \textbf{0.499} &\textbf{0.661} &0.960 &\textbf{0.687} &0.977\\ 
\bottomrule       
\end{tabular}
}
\caption{Quantitative evaluation of FreeCOS compared with different methods on the XCAD dataset.}\label{XCAD_sota}
\end{table}

\begin{table}[]
\centering
\resizebox{1\linewidth}{!}
{
\begin{tabular}{cl|cccc|cccc}
\toprule
& \multirow{2}{*}{Methods} & \multicolumn{4}{c|}{DRIVE}   & \multicolumn{4}{c}{STARE} \\ 
& & Acc. & Sn. & Sp. & AUC & Acc. & Sn. & Sp. &AUC \\
\midrule   
\multirow{4}{*}{Traditional} 
    &Hessian~\cite{frangi1998multiscale} &0.941	&0.644	&0.97	&0.847	&0.938	&0.690	&0.957	&0.858\\
    &OOF~\cite{law2008three} &0.936	&0.688	&0.959	&0.920	&0.920	&0.770	&0.932	&0.955\\
    &Memari~\cite{memari2019retinal} &\textbf{0.961}	&0.761	&0.981	&0.871	&0.951	&0.782	&0.965	&0.783\\
    &Khan~\cite{khan2020hybrid} &0.958	&0.797	&0.973	&0.885	&\textbf{0.996}	&0.792	&\textbf{0.998}	&0.895
 \\ \midrule
\multirow{2}{*}{Unsupervised} 
    &IIC~\cite{ji2019invariant} &0.738	&0.632	&0.840	&0.736	&0.710	&0.586	&0.832	&0.709\\
    &ReDO~\cite{chen2019unsupervised} &0.761	&0.593	&0.927	&0.760	&0.756	&0.567	&0.899	&0.733
 \\ \midrule
 \multirow{3}{*}{Self-supervised} 
    &SSVS~\cite{ma2021self} &0.913	&0.794	&\textbf{0.982}	&0.888	&0.910	&0.774	&0.980	&0.877\\
    &DARL~\cite{kim2022diffusion} &--	&0.456	&--	&--	&--	&0.480	&--	&--\\    
    &Ours &0.921	&\textbf{0.810}	&0.932	&\textbf{0.941}	&0.952	&\textbf{0.797}	&0.964	&\textbf{0.971}
 \\ \bottomrule
\end{tabular}
}
\caption{Quantitative evaluation of FreeCOS compared with different methods on the retinal dataset.}
\label{Retinal_sota}
\end{table}

\begin{table}[]
\centering
\resizebox{1\linewidth}{!}{
\begin{tabular}{clccccc}
\toprule
                             & Methods             & AUC    & Dice      & Acc.      & Sn.      & Sp.\\ \hline
\multirow{2}{*}{Traditional}  
                                          & Hessian~\cite{frangi1998multiscale} &0.780 &0.122	&0.935	&0.310	&0.945\\
                                          & OOF~\cite{law2008three}  &0.482	&0.031	&0.770	&0.244	&0.778\\ 
                                          \midrule
\multirow{1}{*}{Unsupervised}  
                                          & ReDO~\cite{chen2019unsupervised} &0.422	&0.035	&0.632	&0.450	&0.635\\ 
                                          \midrule       
\multirow{3}{*}{Self-supervised}          & SSVS~\cite{ma2021self}  & 0.477	&0.078	&0.299	& 0.042	&0.912\\
& DARL~\cite{kim2022diffusion}  & 0.888	&0.395	&\textbf{0.974}	& 0.542	&\textbf{0.981}\\
& Ours  &\textbf{0.920}	&\textbf{0.525}	&\textbf{0.974}	&\textbf{0.576}	&0.979\\ \bottomrule          
\end{tabular}
}
\caption{Quantitative evaluation of FreeCOS compared with different methods on the CrackTree dataset.}\label{Crack_sota}
\end{table}

Table.~\ref{XCAD_sota} compares the performance of vessel segmentation on XCAD between FreeCOS and the state-of-the-art methods, including the unsupervised methods~\cite{chen2019unsupervised,ji2019invariant}, the self-supervised methods~\cite{ma2021self,kim2022diffusion}, domain adaptation methods~\cite{bermudez2018domain,roels2019domain} and the traditional methods~\cite{frangi1998multiscale,law2008three}. The results in the 1st row are based on supervised U-Net as~\cite{ma2021self} (i.e., identical segmentation network trained using real images with manual labels) which are the upper bound. 

For domain adaption methods, we pretrain a vessel segmentation model based on U-Net using training images of DRIVE. Then we adapt the pre-trained model to XCAD using  MMD~\cite{bermudez2018domain} and Ynet~\cite{roels2019domain}. Even with supervised information in the annotated source domain, the performance of MMD and YNet is still inferior to ours. Specifically, our method achieves 21.2\% improvement in Jaccard, 22.7\% improvement in Dice, 6.9\% improvement in Acc, 16.4\% improvement in Sn, and 4.2\% improvement in Sp compared with YNet. 

Compared with the unsupervised methods IIC~\cite{ji2019invariant} and ReDO~\cite{chen2019unsupervised}, our method achieves significantly better performance for all metrics on XCAD. The results show that unsupervised methods cannot achieve satisfactory performance on the gray-scale X-ray images where the segmentation objects can be hardly distinguished from the background. 

Self-supervised method SSVS~\cite{ma2021self} is specifically designed for XCAD, yet our method still achieves much better performance for all metrics, i.e., 11\% improvement in Jaccard, 10.4\% improvement in Dice and 10.4\% improvement in Sn.

\begin{figure*}
\begin{center}
\includegraphics[width=0.9\linewidth]{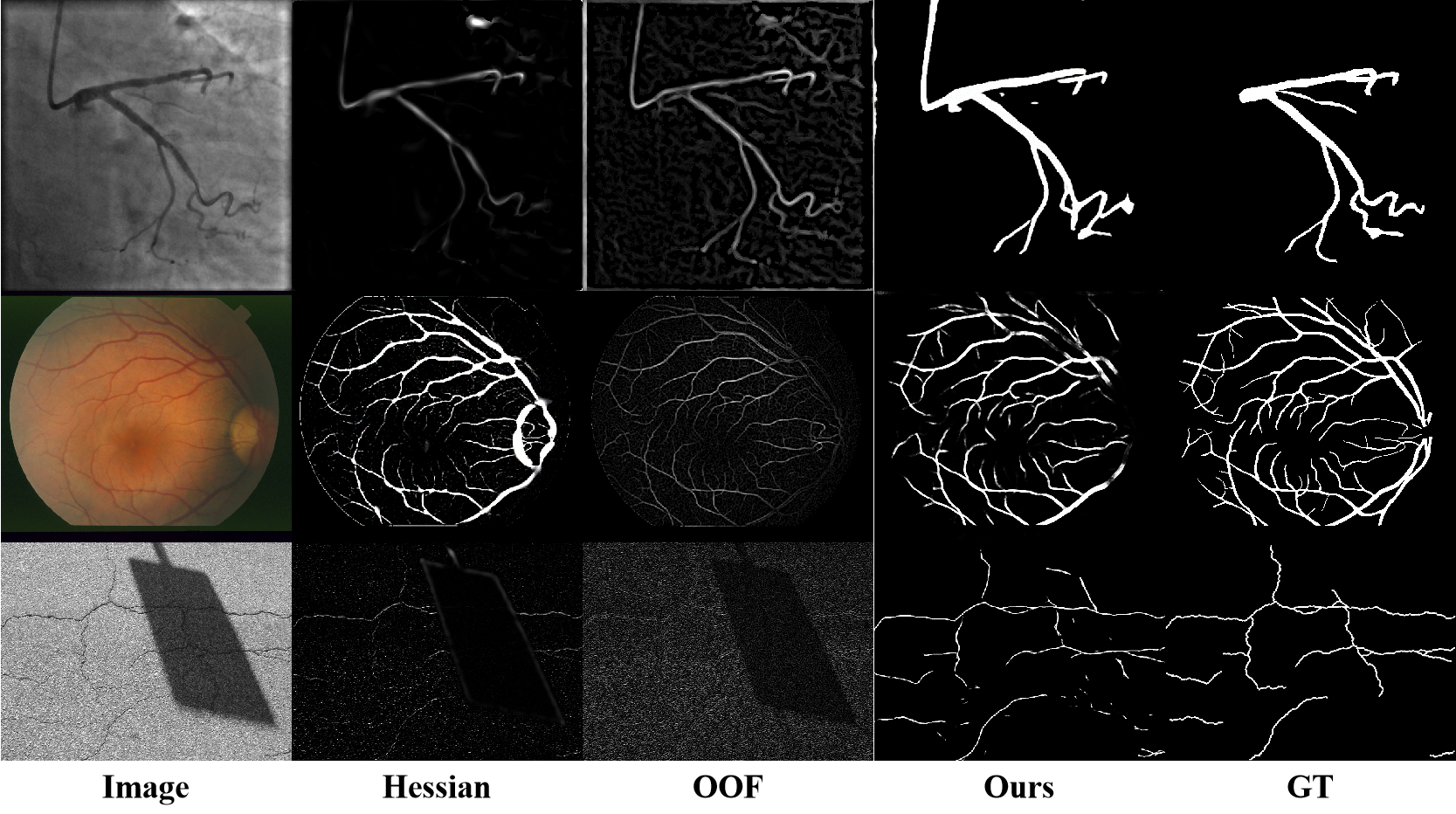}
\end{center}
\caption{The visualization results of images from the various kinds of curvilinear datasets.}
\label{fig:result}
\vspace{-7pt}
\end{figure*}

Table.~\ref{Retinal_sota} and~\ref{Crack_sota} further compare our method with the existing methods on the retinal and crack datasets and a similar trend can be observed in these three datasets. Figure.~\ref{fig:result} shows the visualization results of images from various kinds of curvilinear datasets.
\subsubsection{Ablation Study}
\begin{table}
\centering
\resizebox{1\linewidth}{!}{
\begin{tabular}{lccccc}
\toprule
             & Jaccard    & Dice      & Acc.      & Sn.      & Sp.\\
                             \midrule
\multirow{1}{*}{FFS} &0.450	&0.615	&0.955	&0.664	&0.974\\
                                          \midrule
\multirow{1}{*}{FFS+PSAL}  &0.468	&0.633	&0.957	&0.680	&0.974\\
                                          \midrule
\multirow{1}{*}{FFS+PSAL+IA}  &0.485	&0.647	&0.958	&0.667	&0.976\\
                                          \midrule            
\multirow{1}{*}{FFS+GIA}  &\textbf{0.499}	&\textbf{0.661}	&\textbf{0.960}	&\textbf{0.687}	&\textbf{0.977}\\ 
                                          \bottomrule       
\end{tabular}
}
\caption{Ablation study for modules.}\label{tab:ablationmodule}
\end{table}

\begin{table}
\centering
\resizebox{1\linewidth}{!}{
\begin{tabular}{lccccc}
\toprule
             & Jaccard    & Dice      & Acc.      & Sn.      & Sp.\\
                             \midrule
FFS w/o Gaussian blur   &0.383	&0.547	&0.940	&0.654	&0.959\\
                                          \midrule
\multirow{1}{*}{FFS w/o FDA} &0.302	&0.459	&0.920	&0.609	&0.940\\
                                          \midrule
\multirow{1}{*}{FFS w/o various intensities} &0.405	&0.569	&\textbf{0.957}	&0.525	&\textbf{0.984}\\
                                          \midrule
\multirow{1}{*}{FFS w/o various angles}  &0.267	&0.415	&0.939	&0.409	&0.972\\
                                          \midrule
\multirow{1}{*}{FFS w/o various lengths}  &0.224	&0.358	&0.943	&0.300	&0.983\\
                                          \midrule
\multirow{1}{*}{FFS w/o various widths}  &0.224	&0.354	&0.935	&0.348	&0.972\\
                                          \midrule
\multirow{1}{*}{FFS}  &\textbf{0.450}	&\textbf{0.615}	&0.955	&\textbf{0.664}	&0.974\\  
                                          \bottomrule       
\end{tabular}
}
\caption{Ablation study for FFS.}\label{tab:ablationFFS}
\vspace{-7pt}
\end{table}

\begin{table}
\centering
\resizebox{1\linewidth}{!}{
\begin{tabular}{cccccc}
\toprule
Numbers of images              & Jaccard    & Dice      & Acc.      & Sn.      & Sp.\\
                             \midrule
Synthetic-15  &0.464	&0.629	&0.958	&0.630	&0.979\\
                                          \midrule
\multirow{1}{*}{Synthetic-45} &0.478	&0.642	&0.958	&0.674	&0.976\\
                                          \midrule
\multirow{1}{*}{Synthetic-75}  &0.488	&0.652	&\textbf{0.960}	&0.668	&\textbf{0.978}\\
                                          \midrule
\multirow{1}{*}{Synthetic-150} &\textbf{0.499}	&\textbf{0.661}	&\textbf{0.960}	&\textbf{0.687}	&0.977\\
                                          \bottomrule       
\end{tabular}
}
\caption{Ablation study for different numbers of synthetic images.}\label{tab:ablationfake}
\end{table}

\begin{table}
\centering
\resizebox{1\linewidth}{!}{
\begin{tabular}{cccccc}
\toprule
Numbers of images             & Jaccard    & Dice      & Acc.      & Sn.      & Sp.\\
                             \midrule
Target-162  &0.455	&0.620	&0.952	&0.678	&0.969\\
                                          \midrule
\multirow{1}{*}{Target-486} &0.476	&0.637	&0.955	&0.683	&0.972\\
                                          \midrule
\multirow{1}{*}{Target-810}  &0.466	&0.630	&0.958	&0.635	&\textbf{0.978}\\
                                          \midrule
\multirow{1}{*}{Target-1620} &\textbf{0.499}	&\textbf{0.661}	&\textbf{0.960}	&\textbf{0.687}	&0.977\\
                                          \bottomrule       
\end{tabular}
}
\caption{Ablation study for different numbers of target images.}\label{tab:ablationtarget}
\vspace{-7pt}
\end{table}

We first conduct ablation studies to evaluate the impact of different modules. To this end, we build the following variants based on our method. For all the variant models and our final model, we use the same U-Net model as our backbone. 1) \textbf{FFS}. We utilize the FFS to generate synthetic training images and the corresponding labels to train the U-Net segmentation model. 2) \textbf{FFS+PSAL}. We further apply PSAL to align features of synthetic and real target images on top of FFS. 3) \textbf{FFS+PSAL+IA}. We apply image-level alignment via relative intensity order transformation (i.e., IA) on top of \textbf{FFS+PSAL}. 4) \textbf{FFS+GIA}. We apply our GIA module (including IA, PSAL and CMCL) on top of FFS. This model is also our final curvilinear segmentation model. The ablation studies are conducted on XCAD. The trend on other datasets are similar and thus the results on the other datasets are omitted due to the space limit.

The results in Table~\ref{tab:ablationmodule} show that 1) by training a U-Net model using synthetic images from FFS, we can already achieve better performance than the SOTA self-supervised method SSVS which is based on adversarial learning~\cite{ma2021self}. Such results reveal a very interesting phenomenon although adversarial learning can synthesize visually similar images as real target images, it cannot explicitly control the generated visual patterns and thus fail to enforce the segmentation-oriented patterns and properties to be encoded in synthetic images. In comparison, FFS can explicitly control the synthesis of both curvilinear structures and background patterns and hence can achieve better performance. 2) Aligning features via prediction space adaptation loss (PSAL) can help reduce domain shifts between synthetic images and real target images and thus \textbf{FFS+PSAL} achieves performance improvements compared with FFS. 3) Aligning synthetic images and real target images via IA method can also reduce the domain shifts and thus \textbf{FFS+PSAL+IA} can provide complementary improvements to \textbf{FFS+PSAL}. 4) Finally, the best performance is obtained when combining both FFS and GIA.

We further explore the importance of different parameters in FFS on the final performance and discuss how to generate images of curvilinear structures to encode sufficient and comprehensive visual cues for learning robust and distinctive features. To this end, we build 6 variant models based on FFS. 1) \textbf{FFS w/o Gaussian blur}. We do not perform Gaussian blur to synthetic images from FFS. 2) \textbf{FFS w/o FDA}. We remove FDA-based synthesis from FFS and only generate curvilinear structures without backgrounds. 3) \textbf{FFS w/o various intensities}. We remove intensity variations $c_i$ in the Fractal system and set the intensity to a fixed value 60. 4) \textbf{FFS w/o various angles}. We utilize a small branching angle range from $1^{\circ}$ to $5^{\circ}$ to reduce the branching angle variations. 5) \textbf{FFS w/o various lengths}. We reduce the length variation range to (10px, 20px). 6) \textbf{FFS w/o various widths}. We reduce the width variation to (1px, 5px). 

The results in Table~\ref{tab:ablationFFS} show that 1) Gaussian blur can smooth the transition region between curvilinear objects and backgrounds, better mimicking the target images and providing more challenging samples for training than those with sharp boundaries. As a result, excluding Gaussian blur decreases the performance of FFS by 6.8\% in Dice. 2) FDA could provide background patterns of target images which are essential for learning features of negative samples. Thus, \textbf{FFS w/o FDA} is 15.6\% worse than FFS in Dice. 3) Among the four parameters of the Fractal system, i.e., intensity, angle, length and width, width plays the most important role in the final performance, i.e., \textbf{FFS w/o various widths} decreases the performance by 26.1\% compared with FFS in Dice. To summarize, we identify that the appearance patterns in the curvilinear object, object-background transition regions and background regions are important for self-supervised segmentation. Meanwhile, proper parameter settings which can provide similar geometric characteristics with the real target images are key to the success of our self-supervised method.

\subsubsection{Self-supervised training with more data}
We also examine the segmentation performance when varying the number of synthetic images on the XCAD dataset. Results in Tables.~\ref{tab:ablationfake} and~\ref{tab:ablationtarget} provide the results when using different numbers of synthetic structures and real target images for generating synthetic images for self-supervised training respectively. Synthetic-15 $\sim$ Synthetic-150 denote using the number of synthetic fractal object images $X_{frac}$ from 15 to 150, and Target-162 $\sim$ Target-1620 denote using the number of target images $X_{t}$ from 162 to 1620. Increasing both synthetic structures and real target images can accordingly improve the final segmentation performance. 

\section{Conclusion}
\label{sec:con}
In this paper, we propose a novel self-supervised curvilinear object segmentation method that learns robust and distinctive features from fractals and unlabeled images (FreeCOS).
Different from existing methods, FreeCOS applies the proposed FFS and GIA approach to effectively guide learning distinctive features to distinguish curvilinear objects and backgrounds and aligns information of synthetic and target images at both image and feature levels.
One limitation of FreeCOS is that may generate false positives in other curvilinear objects (e.g., catheters in XCAD) and requires proper selection for width range. However, we have successfully utilized this self-supervised learning method for coronary vessel segmentation, retinal vessel segmentation and crack segmentation by reasonable parameter selection. 
To the best of our knowledge, FreeOCS is the first self-supervised learning method for various curvilinear object segmentation applications.

{\small
\bibliographystyle{ieee_fullname}
\bibliography{iccv2023AuthorKit/arxiv}
}

\end{document}